\documentclass[10pt,twocolumn,letterpaper]{article}

\usepackage[pagenumbers]{cvpr}   % 预印本：显示作者 + 强制页码

\definecolor{cvprblue}{rgb}{0.21,0.49,0.74}
\usepackage[pagebackref,breaklinks,colorlinks,allcolors=cvprblue]{hyperref}
\usepackage{graphicx}
\usepackage{booktabs}
\usepackage{multirow}

\def\confName{CVPR}
\def\confYear{2027}

\title{CorrectCast: Selective Multimodal Context for Scale-Aware Residual Correction in Precipitation Nowcasting}

\author{
Henan Wang \and
Shengwu Xiong \and
Yifang Zhang \and
Pengfei Duan$^\ast$
\\
Wuhan University of Technology
\\
Wuhan, Hubei, China
\\
{\tt\small \{361332, xiongsw, yifangzhang, duanpf\}@wut.edu.cn}
\\
$^\ast$ Corresponding author
}

\begin{document}
\maketitle
\begin{abstract} 
Precipitation nowcasting models suffer from errors in intensity, location, and spatial structure. Addressing these errors by retraining nowcaster backbones can be computationally expensive and requires architecture-specific modifications. To overcome these limitations, we propose CorrectCast, a decoupled residual correction framework that refines completed nowcasts while keeping the base nowcaster frozen. By operating on predicted precipitation fields, CorrectCast requires training only a lightweight corrector, reducing computational overhead while providing a unified correction interface across different nowcasting models. Conditioned on the base nowcast and historical quantitative precipitation estimation (QPE) and precipitable water vapor (PWV) observations, the corrector selects multimodal context and employs scale-aware modulation to estimate residual corrections. On the Mississippi River Basin (MRB) and contiguous United States (CONUS) datasets, CorrectCast improves six base nowcasters spanning extrapolation-based, deterministic, and generative approaches, with absolute gains in mean CSI averaging 0.0233 and 0.0290 across the six models, respectively.
\end{abstract}

\section{Introduction}
\label{sec:intro}
Precipitation nowcasting plays a critical role in providing timely warnings for rainfall-related hazards. Weather, climate and water disasters caused over two million deaths and \$4.3 trillion in losses during 1970--2021 \cite{wmo}. Existing nowcasting approaches span extrapolation-based methods \cite{pysteps}, deterministic networks \cite{MMRNN,CM-STJointNet,alphaPre,exprecast}, and generative models \cite{dgmr,nowcastnet,prediff,cascast,wavec2r,mocast}. Despite advances, these approaches remain susceptible to errors in precipitation intensity, location, and spatial structure. Extrapolation-based methods can struggle with precipitation growth and decay, deterministic networks often produce overly smooth nowcasts at longer lead times, and generative models can exhibit spatial misalignment or structural discrepancies. These limitations motivate a unified approach to refining nowcasts across diverse nowcasting architectures.
\begin{figure}[t]
    \centering
    \includegraphics[width=1.0\linewidth]{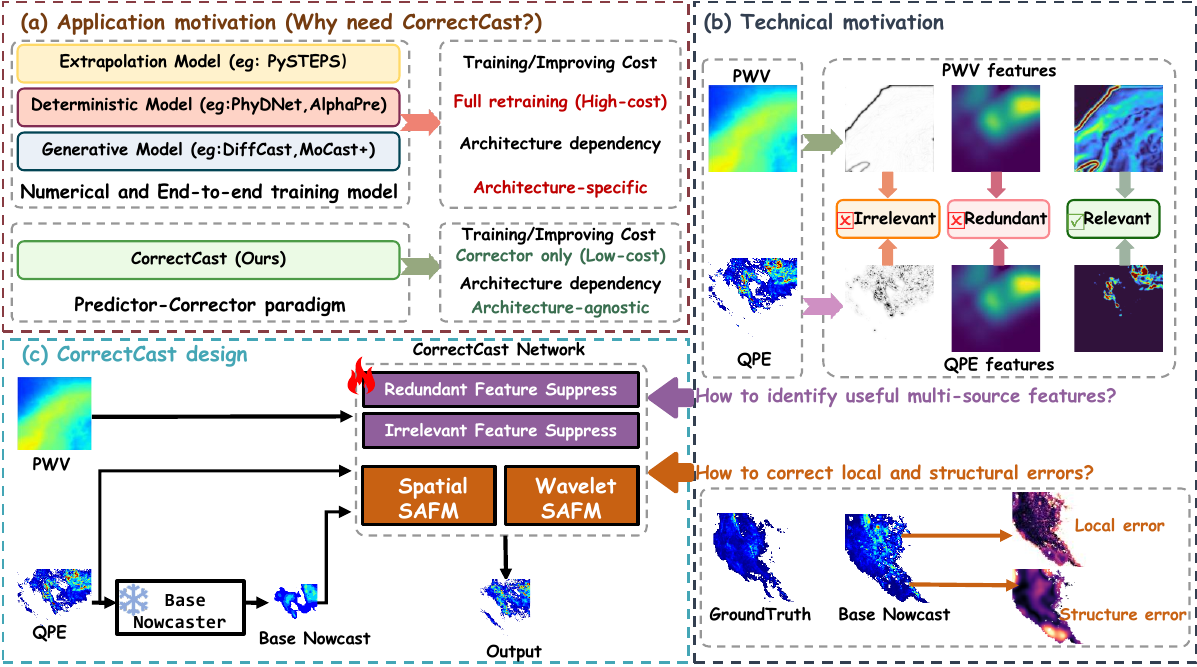}
    \caption{Motivation and architectural design of CorrectCast.}
    \label{fig:different}
\end{figure}

As illustrated in Fig.~\ref{fig:different}(a), extrapolation-based methods, deterministic networks, and generative models employ different prediction mechanisms, making internal changes dependent on the specific architecture. Improving their nowcasts may involve revising prediction modules or retraining network backbones. Such strategies can require architecture-specific adaptations and incur substantial computational costs. DiffCast compares end-to-end training with a frozen-backbone predictor-corrector variant, reporting higher CSI under joint training for three deterministic backbones on SEVIR \cite{diffcast}. However, this result does not establish whether output-level correction can work effectively across heterogeneous, frozen base nowcasters. We therefore investigate a complementary approach: a decoupled predictor-corrector framework that learns residual corrections from completed nowcasts without accessing internal model features or updating the base nowcaster.

However, learning effective residual corrections from completed nowcasts poses two key challenges, as illustrated in Fig.~\ref{fig:different}(b). First, the corrector must identify historical information that is relevant to errors remaining in the base nowcast. Historical QPE captures recent precipitation patterns, while PWV provides complementary information on atmospheric moisture conditions \cite{qpe,qpe2,pwvrelationship,liu2025deep}. However, some QPE information may already be encoded in the base nowcast, and PWV features may include both precipitation-relevant variations and signals unrelated to residual errors. Directly combining these observations may therefore introduce redundancy and task-irrelevant information, limiting their value for correction. Second, the corrector must account for errors at different spatial scales. Local discrepancies call for flexible, position-dependent adjustments across neighboring locations, whereas broader structural errors involve the extent and organization of precipitation systems. Addressing these larger-scale discrepancies requires coordinated adjustments across spatially related regions, rather than independent corrections at individual locations.

To address these challenges, we propose CorrectCast, a decoupled residual correction framework that integrates selective multimodal context construction with scale-aware feature modulation (SAFM), as illustrated in Fig.~\ref{fig:different}(c). Given a completed nowcast from a frozen base nowcaster, CorrectCast estimates a residual correction conditioned on historical QPE and PWV observations. To construct informative context, we extract multiscale precipitation features from QPE and characterize PWV in terms of moisture state, temporal variation, and spatial distribution. We then employ a two-stage information selection strategy: cross-modal orthogonal decomposition suppresses PWV components linearly aligned with QPE features, and a spatiotemporal variational information bottleneck (VIB) subsequently constrains the information retained in the fused representation. To address errors at different spatial scales, Spatial SAFM enables flexible, position-dependent modulation of nowcast features, while Wavelet SAFM applies low-frequency wavelet modulation to encourage coordinated adjustments across broader regions. The resulting features are decoded into a residual correction that refines the original nowcast.

Our main contributions are summarized as follows:
\begin{itemize}
\item We introduce CorrectCast, an output-level correction framework that refines completed nowcasts without modifying base nowcasters or accessing their internal features, providing a common correction interface across different nowcasting paradigms.
\item We develop a correction-oriented representation strategy that combines precipitation and moisture information through cross-modal orthogonal decomposition and a spatiotemporal VIB to reduce redundancy and encourage task-relevant context.
\item We design Spatial SAFM and Wavelet SAFM to support flexible local adjustments and coherent large-scale structural refinement, and demonstrate the effectiveness of CorrectCast across six base nowcasters on two datasets with different geographic coverage.
\end{itemize}

\section{Related Work}
\subsection{Precipitation Nowcasting Paradigms}
Precipitation nowcasting spans numerical extrapolation, deterministic, and generative methods. Numerical methods such as PySTEPS \cite{pysteps} efficiently advect observed precipitation but struggle with growth and decay. Deterministic models learn precipitation evolution from historical observations but may produce smooth nowcasts at longer lead times \cite{convlstm,simVP,earthformer,phydnet,TrajGRU,PredRNN,PredRNN++}. Generative models, including GAN-based \cite{dgmr,nowcastnet,Rsggan} and diffusion-based approaches \cite{prediff,diffcast}, can preserve fine-scale structures but may exhibit spatial displacement or structural discrepancies \cite{nowcastnet,diffcast}.

\subsection{Generative Refinement and Output Correction}
DiffCast combines a deterministic nowcasting backbone with a residual diffusion model, favoring end-to-end optimization while also evaluating a frozen backbone predictor-corrector variant \cite{diffcast}. MoCast+ extends the physics-guided MoCast predictor with diffusion-based residual post-processing \cite{mocast}. These approaches demonstrate the value of residual refinement but primarily study diffusion-based residual refinement within specific nowcasting frameworks. WaveC2R employs wavelet-guided diffusion to refine satellite radar retrievals, demonstrating the potential of refinement with frequency awareness for precipitation structures \cite{wavec2r}. CorrectCast instead investigates residual correction as an independent stage applied to completed nowcasts from heterogeneous base models. CorrectCast trains a lightweight corrector conditioned on the base nowcast and historical QPE and PWV, without accessing internal model features or updating the base nowcaster. By separating correction from prediction, CorrectCast provides a common output-level interface across nowcasting paradigms, with additional training restricted to the corrector.
\begin{figure*}[t]
\centering
\includegraphics[width=\textwidth]{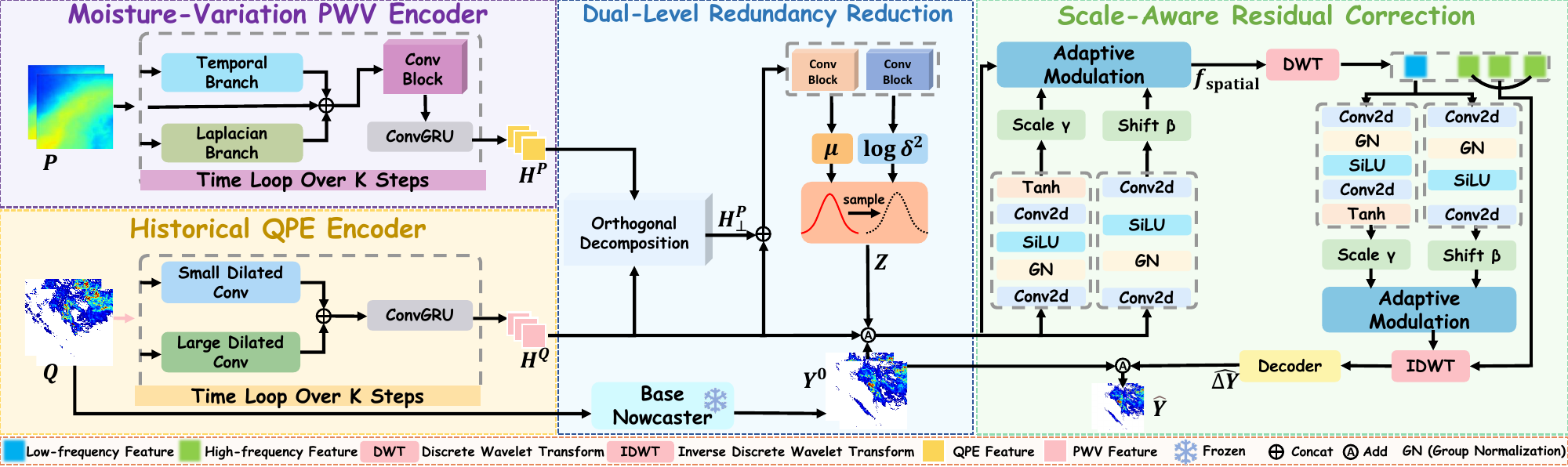}
\caption{Overall architecture of the proposed CorrectCast framework.}
\label{fig:model_structure}
\end{figure*}

\section{Problem Formulation}
\label{sec:problem}
Let $Y^{0}\in\mathbb{R}^{T\times H\times W}$ denote a completed nowcast produced by a base nowcaster, and let $Y\in\mathbb{R}^{T\times H\times W}$ denote the corresponding ground-truth future precipitation sequence. Historical QPE and PWV observations are denoted by $Q\in\mathbb{R}^{K\times H\times W}$ and $P\in\mathbb{R}^{K\times H\times W}$, respectively. Here, $K$ and $T$ denote the numbers of historical frames and nowcast frames, while $H$ and $W$ represent the spatial dimensions. CorrectCast predicts a residual correction:
\begin{equation}
    \widehat{\Delta Y}=\mathcal{G}_{\Theta}\left(Y^{0},Q,P\right),
    \label{eq:residual_function}
\end{equation}
where $\mathcal{G}_{\Theta}$ denotes the learnable corrector parameterized by $\Theta$, and the target residual is defined as $\Delta Y=Y-Y^{0}$. The corrected nowcast is then obtained as
\begin{equation}
    \widehat{Y}=\Pi_{+}\left(Y^{0}+\widehat{\Delta Y}\right),
    \label{eq:refined_prediction}
\end{equation}
where $\Pi_{+}(\cdot)$ denotes element-wise projection onto the nonnegative domain. Throughout training, the base nowcaster remains fixed, and only the corrector parameters $\theta$ are optimized. CorrectCast thus learns to refine completed nowcasts using historical multimodal observations without updating the base nowcaster.

\section{Methodology}
\subsection{Overview}
CorrectCast consists of two stages: selective multimodal context construction and scale-aware residual correction. As illustrated in Fig.~\ref{fig:model_structure}, historical QPE and PWV are first encoded into candidate correction features. Cross-modal orthogonal decomposition first suppresses PWV components linearly aligned with QPE representations, followed by a spatiotemporal VIB that encourages a compact fused representation with reduced input dependence while preserving information relevant to residual correction. Guided by this context, Spatial SAFM applies flexible, position-dependent modulation to the completed nowcast, while Wavelet SAFM subsequently introduces low-frequency wavelet modulation to support coordinated refinement of broader precipitation structures. Finally, a decoder predicts the residual correction $\widehat{\Delta Y}$, which is combined with $Y^{0}$ according to Eq.~\eqref{eq:refined_prediction}.

\subsection{Selective Multimodal Context Construction}
\subsubsection{Historical QPE Encoder}
The base nowcaster already uses historical QPE observations to model the evolution of the precipitation field. Rather than repeating the full nowcasting process, the QPE branch of CorrectCast extracts compact multiscale features to support residual correction. Given a historical QPE sequence $Q=\{Q_t\}_{t=1}^{K}$, dilated convolutions $\phi_s(\cdot)$ are applied to each frame $Q_t$ with dilation rates $\mathcal{S}=\{2,4\}$:
\begin{equation}
    F_t^{(s)}=\phi_s(Q_t), \qquad s\in\mathcal{S}.
\end{equation}
The resulting features are concatenated to form the multiscale representation of each frame:
\begin{equation}
    F_t=\operatorname{Concat}_{s\in\mathcal{S}}\left(F_t^{(s)}\right).
\end{equation}
A ConvGRU then captures temporal dependencies across $\{F_t\}_{t=1}^{K}$ to produce the QPE representation $H^{Q}$.

\subsubsection{Moisture-Variation PWV Encoder}
PWV provides information about atmospheric moisture conditions  that is complementary to the historical QPE. To characterize the moisture state and its temporal and spatial variations, each PWV frame $P_t$ is augmented with a temporal-difference descriptor $D_t^{\mathrm{td}}$ and a spatial-Laplacian descriptor $D_t^{\mathrm{sl}}$. We further apply spatial normalization to the PWV field and its derived descriptors. The resulting frame-level input is
\begin{equation}
    G_t=\operatorname{Concat}\left(P_t, \mathcal{M}(P_t),
    \mathcal{M}(D_t^{\mathrm{td}}),
    \mathcal{M}(D_t^{\mathrm{sl}})
    \right),
    \label{eq:pwv_representation}
\end{equation}
where $\mathcal{M}(\cdot)$ denotes spatial normalization. A convolutional block first transforms $G_t$ into a learned feature representation $E_t$. A ConvGRU then captures temporal dependencies across $\{E_t\}_{t=1}^{K}$, producing the PWV representation $H^{P}$ from its final hidden state.

\subsubsection{Cross-Modal Linear Redundancy Reduction}
QPE and PWV representations may contain overlapping information, and directly retaining such information can introduce redundancy into the residual correction process. To reduce this overlap, we suppress the PWV components linearly aligned with the corresponding QPE features. Specifically, for each sample and channel, we flatten the spatial dimensions of $H^{Q}$ and $H^{P}$ into vectors $h^{Q},h^{P}\in\mathbb{R}^{HW}$. The projection of $h^{P}$ onto $h^{Q}$ is defined as
\begin{equation}
    h_{\parallel}^{P}=
    \frac{\left\langle h^{P},h^{Q}\right\rangle}
    {\left\|h^{Q}\right\|_{2}^{2}+\varepsilon}h^{Q},
    \label{eq:orth_projection}
\end{equation}
where $\varepsilon>0$ ensures numerical stability. Subtracting this projection yields the remaining PWV component $h_{\perp}^{P}=h^{P}-h_{\parallel}^{P}$, which is approximately orthogonal to $h^{Q}$ when $\varepsilon$ is sufficiently small. After restoring the spatial dimensions, we obtain $H_{\perp}^{P}$ and construct the fused representation $X=\operatorname{Concat}\left(H^{Q},H_{\perp}^{P}\right)$. This operation reduces channel-wise linear alignment between QPE and PWV features without assuming complete statistical independence.

\subsubsection{Task-Level Redundancy Reduction}

Since the fused representation $X$ may contain information irrelevant to residual correction, we employ a spatiotemporal VIB to learn a compact, correction-oriented representation $Z$ through a diagonal gaussian posterior:
\begin{equation}
    q_{\theta}(Z\mid X)
    =
    \mathcal{N}\left(\mu,\operatorname{diag}(\sigma^2)\right),
\end{equation}
where $(\mu,\log\sigma^2)=f_{\theta}(X)$, and \(f_{\theta}\) denotes a learnable convolutional block. The latent representation is sampled using the reparameterization trick:
\begin{equation}
    Z
    =
    \mu+\sigma\odot\epsilon,
    \qquad
    \epsilon\sim\mathcal{N}(0,I).
    \label{eq:vib_reparameterization}
\end{equation}

The information bottleneck objective encourages $Z$ to retain correction-relevant information, while limiting its dependence on $X$:
\begin{equation}
    \max_{\theta}
    \;
    I(Z;\Delta Y\mid Y^0,H^Q)
    -
    \beta I(Z;X),
    \qquad
    \beta>0.
    \label{eq:ib_objective}
\end{equation}
Directly computing $I(Z;X)$ is difficult because the marginal distribution of $q_{\theta}(Z)$ is generally intractable. To obtain a computable alternative, we introduce a standard gaussian prior $p(Z)=\mathcal{N}(0,I)$. By the definition of mutual information, the expected KL divergence satisfies
\begin{equation}
\begin{aligned}
&\mathbb{E}_{X}
\left[
D_{\mathrm{KL}}
\left(
q_{\theta}(Z\mid X)
\,\Vert\,
p(Z)
\right)
\right] \\
&\quad =
I(Z;X)
+
D_{\mathrm{KL}}
\left(
q_{\theta}(Z)
\,\Vert\,
p(Z)
\right).
\end{aligned}
\label{eq:kl_mi_decomposition}
\end{equation}
Since the second term is non-negative, the mutual information is upper-bounded by
\begin{equation}
I(Z;X)
\leq
\mathbb{E}_{X}
\left[
D_{\mathrm{KL}}
\left(
q_{\theta}(Z\mid X)
\,\Vert\,
p(Z)
\right)
\right].
\label{eq:ib_upper_bound}
\end{equation}
This provides an alternative to directly minimizing $I(Z;X)$. However, minimizing the KL divergence alone may also remove information needed for residual correction. Therefore, we supervise $Z$ through the complete downstream correction network, together with $Y^0$ and $H^Q$, to predict the precipitation residual:
\begin{equation}
    \widehat{\Delta Y}
    =
    \mathcal{R}_{\phi}
    \left(
        Y^{0},
        H^{Q},
        Z
    \right),
    \label{eq:vib_residual_prediction}
\end{equation}
where $\mathcal{R}_{\phi}$ comprises scale-aware residual correction and residual decoding. To measure the correction accuracy, we define the residual reconstruction loss as
\begin{equation}
    \mathcal{L}_{\mathrm{rec}}
    =
    \left\|
        \Delta Y-\widehat{\Delta Y}
    \right\|_{2}^{2}.
    \label{eq:vib_reconstruction}
\end{equation}
Assuming a fixed-variance isotropic gaussian residual likelihood, minimizing the expected loss in Eq.~\ref{eq:vib_reconstruction} maximizes a variational lower bound on $I(Z;\Delta Y\mid Y^0,H^Q)$.

Finally, the reconstruction loss and KL regularization are combined to jointly encourage correction accuracy and information compression:
\begin{equation}
    \mathcal{L}_{\mathrm{VIB}}
    =
    \lambda_{\mathrm{mse}}
    \mathcal{L}_{\mathrm{rec}}
    +
    \lambda_{\mathrm{kl}}
    D_{\mathrm{KL}}
    \left(
        q_{\theta}(Z\mid X)
        \,\Vert\,
        p(Z)
    \right),
    \label{eq:vib_loss}
\end{equation}
where $\lambda_{\mathrm{mse}}$ and $\lambda_{\mathrm{kl}}$ balance residual fidelity and latent information compression, respectively.

\subsection{Scale-Aware Residual Correction} 
\subsubsection{Spatial SAFM for Local Adaptation}
Let $f_{\mathrm{prior}}$ denote features extracted from the completed base nowcast $Y^{0}$. We first combine the historical QPE representation $H^{Q}$ with the VIB latent representation $Z$ to construct a fused correction context, which is then injected into the nowcast features:
\begin{equation}
Z_A = H^{Q}+\mathcal{J}(Z),\qquad
f_{\mathrm{init}} = f_{\mathrm{prior}}+Z_A,
\end{equation}
where $\mathcal{J}$ denotes a learned projection with feature scaling.

Guided by $Z_A$, Spatial SAFM generates spatially varying scale and shift parameters to refine the initialized features:
\begin{equation}
\begin{aligned}
\gamma_{\mathrm{spatial}} &= g_{\gamma}(Z_A),\qquad
\beta_{\mathrm{spatial}} = g_{\beta}(Z_A),\\
f_{\mathrm{spatial}} &=
f_{\mathrm{init}}\odot(1+\gamma_{\mathrm{spatial}})
+\beta_{\mathrm{spatial}}.
\end{aligned}
\end{equation}
By varying the modulation parameters across spatial locations, Spatial SAFM enables localized enhancement or suppression of feature responses, allowing different adjustments at neighboring positions.
\begin{table*}[t]
\centering
\caption{Quantitative results on the MRB and CONUS datasets. Within each base nowcaster pair, \textbf{boldface} indicates the better value. For LPIPS and MSE, the improvement row reports the absolute reduction.}
\label{tab:merged_mrb_conus_metrics}
\resizebox{\textwidth}{!}{
\begin{tabular}{l|ccccccc|ccccccc}
\toprule
\multirow{3}{*}{Models}
& \multicolumn{7}{c|}{MRB Dataset}
& \multicolumn{7}{c}{CONUS Dataset} \\
\cmidrule(lr){2-8}
\cmidrule(lr){9-15}
& CSI $\uparrow$ & HSS $\uparrow$ & CSI-pool4 $\uparrow$ & CSI-pool16 $\uparrow$ & SSIM $\uparrow$ & LPIPS $\downarrow$ & MSE $\downarrow$
& CSI $\uparrow$ & HSS $\uparrow$ & CSI-pool4 $\uparrow$ & CSI-pool16 $\uparrow$ & SSIM $\uparrow$ & LPIPS $\downarrow$ & MSE $\downarrow$ \\
\midrule
PySTEPS\cite{pysteps}
& 0.2246 & 0.3464 & 0.3332 & 0.5405 & 0.9397 & 0.0606 & 22.6499
& 0.0823 & 0.1358 & 0.1537 & 0.2890 & 0.9810 & 0.0228 & 0.9644 \\
PySTEPS + CorrectCast
& \textbf{0.2406} & \textbf{0.3651} & \textbf{0.3639} & \textbf{0.5731} & \textbf{0.9478} & \textbf{0.0470} & \textbf{18.4720}
& \textbf{0.0953} & \textbf{0.1545} & \textbf{0.1705} & \textbf{0.3315} & \textbf{0.9848} & \textbf{0.0188} & \textbf{0.5464} \\
\emph{Improvement}
& \emph{\textbf{+0.0160}} & \emph{\textbf{+0.0187}} & \emph{\textbf{+0.0307}} & \emph{\textbf{+0.0326}} & \emph{\textbf{+0.0081}} & \emph{\textbf{+0.0136}} & \emph{\textbf{+4.1779}}
& \emph{\textbf{+0.0130}} & \emph{\textbf{+0.0187}} & \emph{\textbf{+0.0168}} & \emph{\textbf{+0.0425}} & \emph{\textbf{+0.0038}} & \emph{\textbf{+0.0040}} & \emph{\textbf{+0.4180}} \\
\midrule
PhyDNet\cite{phydnet}
& 0.3046 & 0.4498 & 0.3293 & 0.3513 & 0.9378 & 0.0943 & 8.7959
& 0.0692 & 0.1157 & 0.0722 & 0.0807 & 0.9852 & 0.0408 & 0.4872 \\
PhyDNet + CorrectCast
& \textbf{0.3222} & \textbf{0.4660} & \textbf{0.3900} & \textbf{0.4817} & \textbf{0.9477} & \textbf{0.0539} & \textbf{8.0004}
& \textbf{0.0733} & \textbf{0.1225} & \textbf{0.0989} & \textbf{0.1604} & \textbf{0.9860} & \textbf{0.0368} & \textbf{0.4596} \\
\emph{Improvement}
& \emph{\textbf{+0.0176}} & \emph{\textbf{+0.0162}} & \emph{\textbf{+0.0607}} & \emph{\textbf{+0.1304}} & \emph{\textbf{+0.0099}} & \emph{\textbf{+0.0404}} & \emph{\textbf{+0.7955}}
& \emph{\textbf{+0.0041}} & \emph{\textbf{+0.0068}} & \emph{\textbf{+0.0267}} & \emph{\textbf{+0.0797}} & \emph{\textbf{+0.0008}} & \emph{\textbf{+0.0040}} & \emph{\textbf{+0.0276}} \\
\midrule
AlphaPre\cite{alphaPre}
& 0.3244 & 0.4748 & 0.3613 & 0.3967 & 0.9500 & 0.0826 & 8.4654
& 0.0883 & 0.1437 & 0.0998 & 0.1150 & 0.9853 & 0.0360 & 0.5328 \\
AlphaPre + CorrectCast
& \textbf{0.3617} & \textbf{0.5167} & \textbf{0.4323} & \textbf{0.5234} & \textbf{0.9570} & \textbf{0.0536} & \textbf{7.6584}
& \textbf{0.1014} & \textbf{0.1637} & \textbf{0.1570} & \textbf{0.2818} & \textbf{0.9858} & \textbf{0.0226} & \textbf{0.4459} \\
\emph{Improvement}
& \emph{\textbf{+0.0373}} & \emph{\textbf{+0.0419}} & \emph{\textbf{+0.0710}} & \emph{\textbf{+0.1267}} & \emph{\textbf{+0.0070}} & \emph{\textbf{+0.0290}} & \emph{\textbf{+0.8070}}
& \emph{\textbf{+0.0131}} & \emph{\textbf{+0.0200}} & \emph{\textbf{+0.0572}} & \emph{\textbf{+0.1668}} & \emph{\textbf{+0.0005}} & \emph{\textbf{+0.0134}} & \emph{\textbf{+0.0869}} \\
\midrule
ExPreCast\cite{exprecast}
& 0.0245 & 0.0456 & 0.0813 & 0.2468 & 0.9312 & 0.0996 & 15.3777
& 0.0072 & 0.0140 & 0.0261 & 0.1069 & 0.9807 & 0.0368 & 0.6544 \\
ExPreCast + CorrectCast
& \textbf{0.0551} & \textbf{0.0946} & \textbf{0.1147} & \textbf{0.2514} & \textbf{0.9429} & \textbf{0.0753} & \textbf{13.0778}
& \textbf{0.0699} & \textbf{0.1149} & \textbf{0.1318} & \textbf{0.2491} & \textbf{0.9850} & \textbf{0.0250} & \textbf{0.5426} \\
\emph{Improvement}
& \emph{\textbf{+0.0306}} & \emph{\textbf{+0.0490}} & \emph{\textbf{+0.0334}} & \emph{\textbf{+0.0046}} & \emph{\textbf{+0.0117}} & \emph{\textbf{+0.0243}} & \emph{\textbf{+2.2999}}
& \emph{\textbf{+0.0627}} & \emph{\textbf{+0.1009}} & \emph{\textbf{+0.1057}} & \emph{\textbf{+0.1422}} & \emph{\textbf{+0.0043}} & \emph{\textbf{+0.0118}} & \emph{\textbf{+0.1118}} \\
\midrule
NowcastNet\cite{nowcastnet}
& 0.1862 & 0.2870 & 0.2601 & 0.3965 & 0.9381 & 0.0732 & 11.5802
& 0.0135 & 0.0260 & 0.0457 & 0.1166 & 0.9321 & 0.1346 & 13.7006 \\
NowcastNet + CorrectCast
& \textbf{0.2158} & \textbf{0.3206} & \textbf{0.2917} & \textbf{0.4150} & \textbf{0.9513} & \textbf{0.0586} & \textbf{9.3595}
& \textbf{0.0902} & \textbf{0.1479} & \textbf{0.1629} & \textbf{0.2517} & \textbf{0.9792} & \textbf{0.0212} & \textbf{0.5790} \\
\emph{Improvement}
& \emph{\textbf{+0.0296}} & \emph{\textbf{+0.0336}} & \emph{\textbf{+0.0316}} & \emph{\textbf{+0.0185}} & \emph{\textbf{+0.0132}} & \emph{\textbf{+0.0146}} & \emph{\textbf{+2.2207}}
& \emph{\textbf{+0.0767}} & \emph{\textbf{+0.1219}} & \emph{\textbf{+0.1172}} & \emph{\textbf{+0.1351}} & \emph{\textbf{+0.0471}} & \emph{\textbf{+0.1134}} & \emph{\textbf{+13.1216}} \\
\midrule
DiffCast\cite{diffcast}
& 0.2822 & 0.4313 & 0.3842 & 0.5338 & 0.9451 & 0.0609 & 12.2483
& 0.0844 & 0.1427 & 0.1532 & 0.2734 & 0.9820 & 0.0287 & 0.7856 \\
DiffCast + CorrectCast
& \textbf{0.2907} & \textbf{0.4392} & \textbf{0.3881} & \textbf{0.5558} & \textbf{0.9557} & \textbf{0.0466} & \textbf{9.0041}
& \textbf{0.0886} & \textbf{0.1453} & \textbf{0.1559} & \textbf{0.2838} & \textbf{0.9846} & \textbf{0.0204} & \textbf{0.5751} \\
\emph{Improvement}
& \emph{\textbf{+0.0085}} & \emph{\textbf{+0.0079}} & \emph{\textbf{+0.0039}} & \emph{\textbf{+0.0220}} & \emph{\textbf{+0.0106}} & \emph{\textbf{+0.0143}} & \emph{\textbf{+3.2442}}
& \emph{\textbf{+0.0042}} & \emph{\textbf{+0.0026}} & \emph{\textbf{+0.0027}} & \emph{\textbf{+0.0104}} & \emph{\textbf{+0.0026}} & \emph{\textbf{+0.0083}} & \emph{\textbf{+0.2105}} \\
\bottomrule
\end{tabular}
}
\end{table*}

\subsubsection{Wavelet SAFM for Coherent Refinement}
To complement local adaptation, Wavelet SAFM applies a discrete wavelet transform to $f_{\mathrm{spatial}}$, separating the low-frequency component $LL$ from high-frequency detail components. Guided by the correction context $Z_A$, the low-frequency component is modulated to support coordinated adjustments across broader spatial regions. The modulated $LL$ and unchanged high-frequency components are then recombined through the inverse discrete wavelet transform, incorporating broader structural adjustments while passing the high-frequency subbands of the spatially modulated features through the wavelet stage unchanged. The reconstructed features are decoded into the residual correction $\widehat{\Delta Y}$, which is combined with $Y^{0}$ and projected onto the nonnegative domain according to Eq.~\eqref{eq:refined_prediction}.

\subsection{Training Objective}
The overall training objective combines the VIB loss in Eq.~\eqref{eq:vib_loss} with two auxiliary losses. A threshold-based weighted binary cross-entropy loss $\mathcal{L}_{\mathrm{bce}}$ encourages accurate detection of high-intensity precipitation, while a spectral-consistency loss $\mathcal{L}_{\mathrm{spec}}$ penalizes differences between the amplitude spectra of the corrected nowcast and ground truth. The total loss is defined as
\begin{equation}
\mathcal{L} = \mathcal{L}_{\mathrm{VIB}}
+ \lambda_{\mathrm{bce}}\mathcal{L}_{\mathrm{bce}}
+ \lambda_{\mathrm{spec}}\mathcal{L}_{\mathrm{spec}}.
\end{equation}

\section{Experiments}
\subsection{Experimental Setup}
\subsubsection{Datasets}
We evaluate CorrectCast on two precipitation nowcasting datasets, MRB and CONUS, which integrate Multi-Radar Multi-Sensor (MRMS) QPE \cite{MRMS} with Global Navigation Satellite System (GNSS)-derived PWV from NGL \cite{pwv}. MRMS QPE is provided at $0.01^\circ$ (~1 km) spatial resolution, and both variables are temporally aligned at 10-min intervals.

MRB: QPE is cropped to $1024 \times 1024$ grids over $32.0^\circ$--$42.23^\circ$N and $93.5^\circ$--$83.25^\circ$W, with PWV from 1,526 quality-controlled GNSS stations bilinearly interpolated onto the same grid. The training, validation, and test periods are Nov. 2020--Jun. 2022, Jul.--Aug. 2022, and Sep. 2022--Aug. 2023, respectively.

CONUS: Following the importance-sampling strategy of DGMR \cite{dgmr}, precipitation events lasting over 4 h are sampled from the full MRMS domain and cropped into $1024 \times 1024$ patches. PWV from 8,229 GNSS stations is bilinearly interpolated and spatially aligned with the QPE patches. The training, validation, and test periods are Nov. 2020--Aug. 2022, Sep.--Oct. 2022, and Nov. 2022--Oct. 2023, respectively.
\subsubsection{Evaluation Protocol}
Following NowcastNet \cite{nowcastnet} and DiffCast \cite{diffcast}, we evaluate categorical nowcast skill using the  critical success index (CSI), heidke skill score (HSS), CSI-pool4, and CSI-pool16 at precipitation thresholds $\tau\in\{16.0,32.0,64.0\}$ mm/h. We additionally report the structural similarity index measure (SSIM), learned perceptual image patch similarity (LPIPS), and mean squared error (MSE) to characterize structural, perceptual, and pixelwise agreement, respectively. The extra details are reported in the supplementary material.
\begin{figure}[h]
    \centering
    \includegraphics[width=1.0\linewidth]{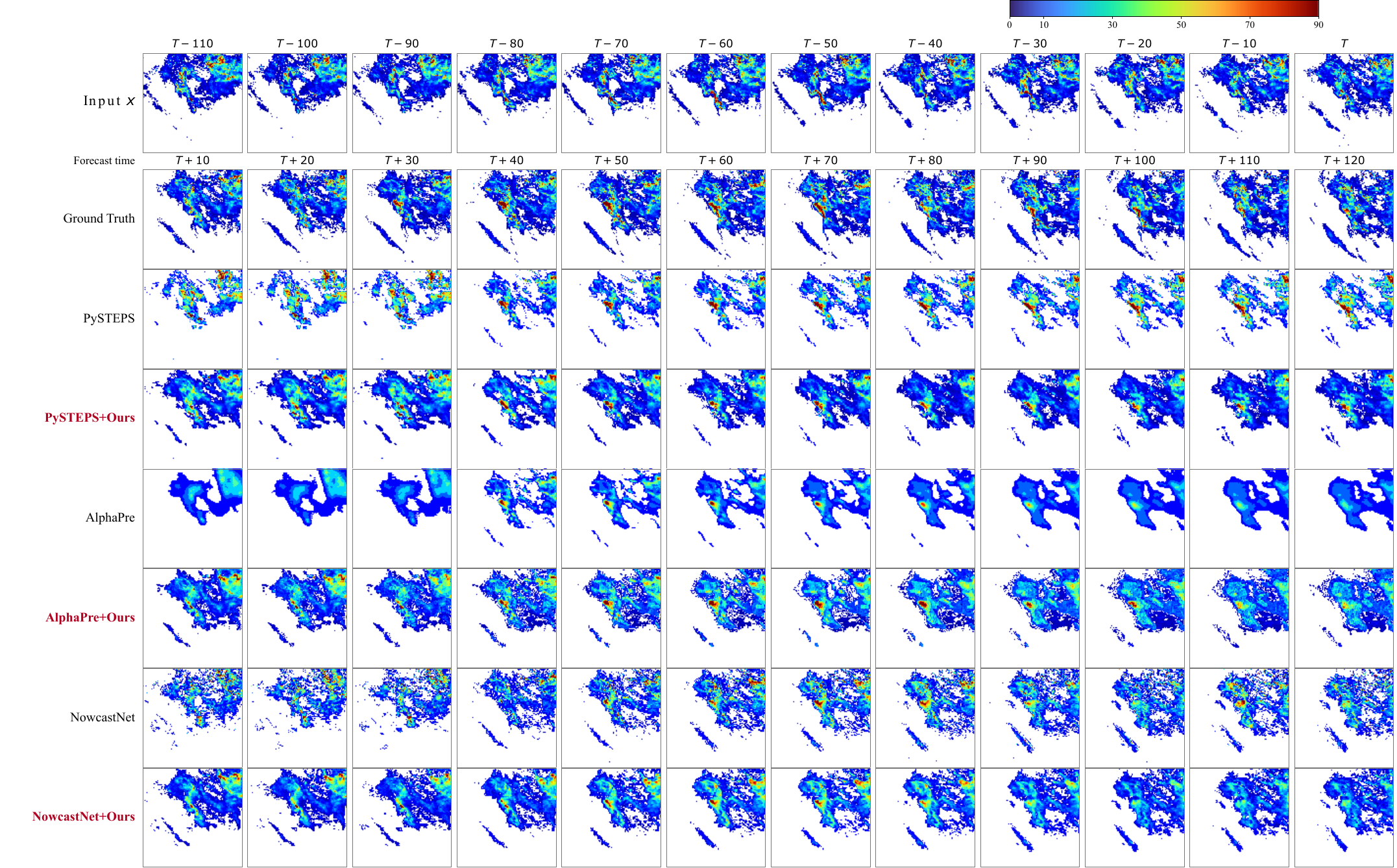}
    \caption{Qualitative comparison for a heavy precipitation event in the MRB dataset. Due to space limitations, the results of PhyDNet, ExPreCast, and DiffCast, together with the CONUS case, are provided in the supplementary material.}
    \label{fig:compare_sample}
\end{figure}
\begin{figure*}[h]
\centering
\includegraphics[width=\textwidth]{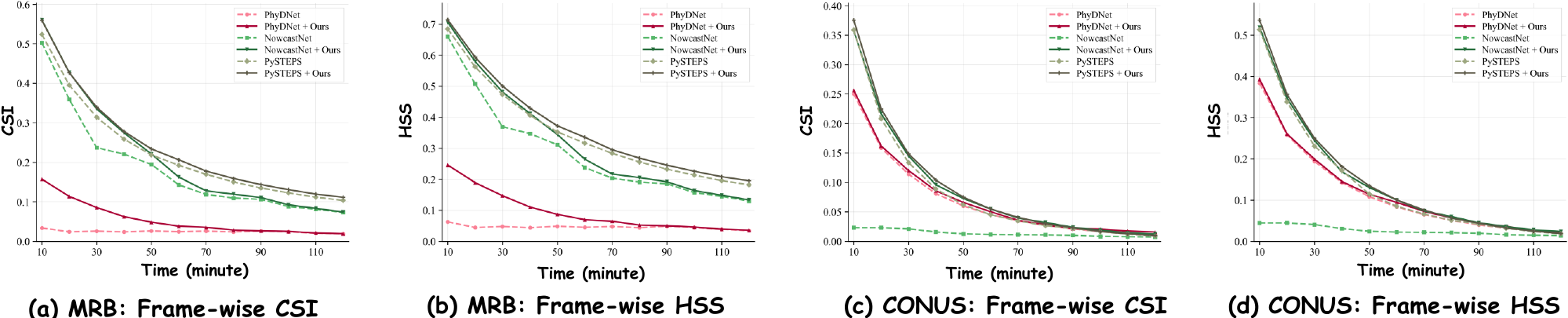}
\caption{CSI and HSS across lead times on the MRB and CONUS datasets. For clarity, only PySTEPS, PhyDNet, and NowcastNet are shown, before and after applying CorrectCast.}
\label{fig:lead_time}
\end{figure*}
\subsection{Experimental Results}
Table~\ref{tab:merged_mrb_conus_metrics} reports the quantitative results on MRB and CONUS. CorrectCast improves CSI and HSS for all six base nowcasters on both datasets, with mean absolute CSI gains of 0.0233 and 0.0290, and corresponding HSS gains of 0.0279 and 0.0452, respectively. On MRB, AlphaPre improves from 0.3244 to 0.3617 in CSI and from 0.4748 to 0.5167 in HSS. The gains also extend to extrapolation-based methods, with PySTEPS achieving a CSI improvement of 0.0160. On CONUS, NowcastNet exhibits particularly large gains, increasing CSI from 0.0135 to 0.0902 and HSS from 0.0260 to 0.1479. 

The improvements extend to spatially pooled and image-quality metrics. Both CSI-pool4 and CSI-pool16 increase for every base nowcaster on the two datasets. Notably, CSI-pool16 improves by 0.1304 for PhyDNet on MRB and 0.1668 for AlphaPre on CONUS. Furthermore, SSIM increases for all six models on both datasets, while LPIPS and MSE consistently decrease. Although the improvement magnitudes vary across models, the consistently positive results support the applicability of CorrectCast across different nowcasting paradigms.
The qualitative comparison in Fig.~\ref{fig:compare_sample} further illustrates the effects of residual correction. The base nowcasts exhibit varying degrees of smoothing, rainband displacement, and local intensity errors. CorrectCast adjusts precipitation intensity and spatial extent while retaining the local structure of precipitation cells.
\begin{table}[t]
\centering
\caption{Quantitative comparison of generative refinement models on the MRB dataset. The best results are highlighted in \textbf{bold} and the second-best results are \underline{underlined}.}
\label{tab:diffcast_phydnet_comparison}
\setlength{\tabcolsep}{3.2pt}
\renewcommand{\arraystretch}{1.05}
\resizebox{\columnwidth}{!}{
\begin{tabular}{l|ccccccc}
\toprule
\multirow{2}{*}{Variants}
& \multicolumn{7}{c}{Metrics} \\
\cmidrule(lr){2-8}
& CSI $\uparrow$ & HSS $\uparrow$ & CSI-pool4 $\uparrow$ & CSI-pool16 $\uparrow$
& SSIM $\uparrow$ & LPIPS $\downarrow$ & MSE $\downarrow$ \\
\midrule

DiffCast(PhyDNet) \cite{diffcast}
& \underline{0.2822} & \underline{0.4313} & \underline{0.3842} & \textbf{0.5338} 
& \underline{0.9451} & 0.0609 & 12.2483 \\
MoCast+(PhyDNet) \cite{mocast}
& 0.2735 & 0.4256 & 0.3705 & 0.4582
& 0.9144 & 0.0789 & 14.3862 \\
WaveC2R \cite{wavec2r}
& 0.2765 & 0.4237 & 0.3639 & 0.4457
& 0.9361 & \underline{0.0600} & \underline{10.0337} \\
CorrectCast(PhyDNet)
& \textbf{0.3222} & \textbf{0.4660} & \textbf{0.3900} & \underline{0.4817} & \textbf{0.9477} & \textbf{0.0539} & \textbf{8.0004} \\
\bottomrule
\end{tabular}
}
\end{table}

Table~\ref{tab:diffcast_phydnet_comparison} compares CorrectCast with representative diffusion-based refinement approaches on MRB. CorrectCast achieves the best results on six of the seven metrics. Compared with DiffCast, it improves CSI by 0.0400 and HSS by 0.0347, while reducing LPIPS from 0.0609 to 0.0539 and MSE from 12.2483 to 8.0004. DiffCast performs better on CSI-pool16. Overall, CorrectCast performs better under most reported metrics, although DiffCast remains superior at the largest pooling scale. Details of the comparison setup are provided in the supplementary material.

Fig.~\ref{fig:lead_time} reports CSI and HSS as functions of nowcast lead time. Both metrics generally decrease at longer horizons, reflecting the increasing difficulty of precipitation nowcasting. CorrectCast nevertheless improves the corresponding base nowcasts over most evaluated lead times. The benefit is therefore not restricted to the first few nowcast frames.
\subsection{Analysis and Discussions}
\subsubsection{Which Components Contribute to CorrectCast?}
\begin{table}[t]
\centering
\caption{
Component ablation of CorrectCast on MRB using NowcastNet as the base nowcaster.
All learned variants use the base nowcast, historical QPE, and moisture-guided PWV as inputs.
Best results are highlighted in \textbf{bold}.
}
\label{tab:ablation_study}
\setlength{\tabcolsep}{2pt}
\renewcommand{\arraystretch}{1.05}
\scriptsize
\resizebox{\columnwidth}{!}{
\begin{tabular}{l|ccccccc}
\toprule
Variants
& CSI $\uparrow$
& HSS $\uparrow$
& CSI-pool4 $\uparrow$
& CSI-pool16 $\uparrow$
& SSIM $\uparrow$
& LPIPS $\downarrow$
& MSE $\downarrow$ \\
\midrule
Base nowcast (NowcastNet)
& 0.1862 & 0.2870 & 0.2601 & 0.3965
& 0.9381 & 0.0732 & 11.5802 \\

Orthogonal decomposition
& 0.1995 & 0.3001 & 0.2739 & 0.4002
& 0.9426 & 0.0683 & 11.0538 \\

Spatiotemporal VIB
& 0.2028 & 0.3032 & 0.2795 & 0.4063
& 0.9432 & 0.0676 & 11.0134 \\

Orthogonal decomposition + VIB
& 0.2060 & 0.3076 & 0.2835 & 0.4101
& 0.9430 & 0.0666 & 10.8972 \\

Orthogonal + VIB + Spatial SAFM
& 0.2074 & 0.3092 & 0.2851 & 0.4111
& 0.9426 & 0.0666 & 10.9269 \\

Orthogonal + VIB + Wavelet SAFM
& 0.2031 & 0.3048 & 0.2797 & 0.4060
& 0.9441 & 0.0672 & 10.9157 \\

Direct prediction (all modules)
& 0.0964 & 0.1559 & 0.1659 & 0.2863
& 0.9367 & 0.0679 & 12.3316 \\

\midrule
\textbf{CorrectCast (all modules)}
& \textbf{0.2158}
& \textbf{0.3206}
& \textbf{0.2917}
& \textbf{0.4150}
& \textbf{0.9513}
& \textbf{0.0586}
& \textbf{9.3595} \\
\bottomrule
\end{tabular}
}
\end{table}
Table~\ref{tab:ablation_study} examines the contributions of information selection, scale-aware modulation, and residual correction. Orthogonal decomposition and spatiotemporal VIB individually improve CSI over the base nowcaster, with their combination further increasing CSI to 0.2060. Adding Spatial SAFM raises CSI to 0.2074, whereas adding Wavelet SAFM alone does not yield an additional gain. Combining both modulation modules achieves the best overall performance, with CSI reaching 0.2158 and HSS reaching 0.3206. Under identical inputs and modules, direct prediction obtains a CSI of 0.0964, substantially below the 0.2158 achieved by residual correction. These results support the combined design of CorrectCast, with additional task and PWV controls reported in the supplementary material.
% Table~\ref{tab:ablation_study} evaluates CorrectCast through three groups of ablations.
% Task controls examine the roles of the base nowcast and historical context. Prior-only refinement (T1) provides little improvement over the base nowcast (T0), while direct nowcasting without the prior (T2) performs poorly. Combining the base nowcast with historical context for residual correction (T3) improves CSI from 0.1862 to 0.2004.

% PWV controls evaluate the contribution of moisture information. Adding CNN-encoded PWV improves CSI from 0.1944 (P0) to 0.2003 (P1), while moisture-guided PWV further increases it to 0.2016 (P2). Under the full configuration, CNN-encoded PWV (P3) achieves 0.2029, compared with 0.2158 using moisture-guided PWV (M7). Temporally shuffling PWV (P4) reduces CSI to 0.2020, highlighting the importance of temporally aligned moisture information.

% Module controls assess the correction mechanisms. Orthogonal decomposition and the VIB together outperform either module individually (M1-M3). Flexible spatial adaptation and coherent wavelet refinement together perform best. Under the same multimodal inputs and correction modules, direct prediction (M6) obtains a CSI of 0.0964, whereas residual correction (M7) reaches 0.2158. This controlled comparison supports the residual formulation used by CorrectCast.
\begin{table}[h]
\centering
\caption{Ablation study of the training losses on the MRB dataset. The best results are highlighted in \textbf{bold} and the second-best results are \underline{underlined}.}
\label{tab:loss_ablation_study}
\setlength{\tabcolsep}{3.2pt}
\renewcommand{\arraystretch}{1.05}
\resizebox{\columnwidth}{!}{
\begin{tabular}{l|ccccccc}
\toprule
\multirow{2}{*}{Variants}
& \multicolumn{7}{c}{Metrics} \\
\cmidrule(lr){2-8}
& CSI $\uparrow$ & HSS $\uparrow$ & CSI-pool4 $\uparrow$ & CSI-pool16 $\uparrow$
& SSIM $\uparrow$ & LPIPS $\downarrow$ & MSE $\downarrow$ \\
\midrule
CorrectCast w/o spectral loss
& 0.2009 & 0.2989 & 0.2557 & 0.3396
& \underline{0.9479} & 0.0755 & \underline{10.4624} \\
CorrectCast w/o BCE loss
& \underline{0.2106} & \underline{0.3129} & \underline{0.2884} & \underline{0.4141}
& 0.9432 & \underline{0.0667} & 10.8713 \\
Full CorrectCast
& \textbf{0.2158} & \textbf{0.3206} & \textbf{0.2917} & \textbf{0.4150}
& \textbf{0.9513} & \textbf{0.0586} & \textbf{9.3595} \\
\bottomrule
\end{tabular}
}
\end{table}
\subsubsection{Training Objective Analysis}
The loss weights are selected by a two-stage validation procedure
(see supplementary material) and then fixed for all experiments. The selected values are
$\lambda_{\mathrm{mse}}=1.0$, $\lambda_{\mathrm{bce}}=0.5$,
$\lambda_{\mathrm{kl}}=10^{-3}$, and $\lambda_{\mathrm{spec}}=0.3$.
Table~\ref{tab:loss_ablation_study} shows that both auxiliary objectives contribute to performance. Removing the spectral loss reduces CSI-pool16 from 0.4150 to 0.3396 and increases LPIPS from 0.0586 to 0.0755, consistent with its role in constraining spatial frequency structure. Removing the BCE term also degrades all reported metrics and increases MSE from 9.3595 to 10.8713. The two terms therefore provide complementary supervision to the residual reconstruction and VIB objectives. 
\begin{figure}[h]
    \centering
    \includegraphics[width=1.0\linewidth]{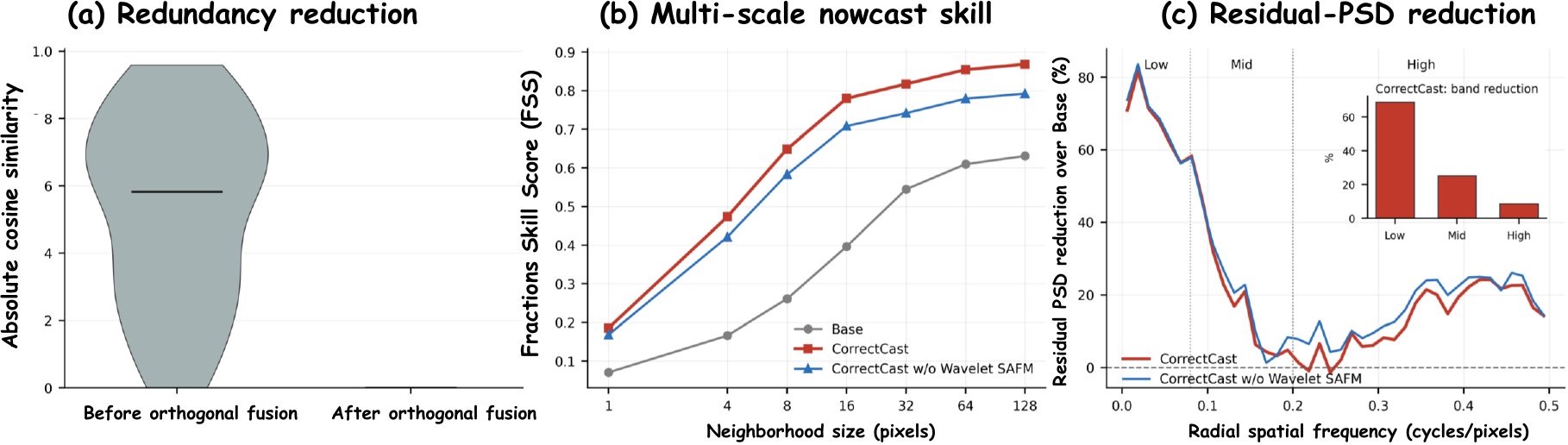}
    \caption{
    Mechanism analysis of CorrectCast.
(a) Absolute cosine similarity between QPE and PWV features before/after orthogonal decomposition.
(b) FSS over neighborhood sizes: base nowcast, CorrectCast, and CorrectCast without Wavelet SAFM.
(c) Residual-PSD reduction over the base nowcast across spatial frequencies; inset gives the mean reduction in low-/mid-/high-frequency bands, positive values indicate lower residual-error.
}
    \label{fig:mechanism}
\end{figure}
\subsubsection{Do the Modules Behave as Intended?}
Fig.~\ref{fig:mechanism}(a) shows that orthogonal decomposition reduces the mean absolute cosine similarity between QPE and PWV features from 0.5299 to $5.97\times10^{-8}$, indicating effective suppression of channel-wise linear alignment. To assess the predictive utility of the retained PWV information, we evaluate residual prediction probes with frozen QPE and PWV encoders. As shown in Table~\ref{tab:orth_probe_main}, adding the QPE-orthogonal component ($H_{\perp}^{P}$) achieves the best CSI, HSS, and CSI-pool16. In contrast, the QPE-aligned component ($H_{\parallel}^{P}$) provides only marginal gains, while including both components reduces performance on these three metrics. However, none of the PWV-augmented probes improves MSE. These results suggest that the retained orthogonal component is particularly useful for improving categorical nowcast skill, although this benefit does not extend to pixelwise MSE in the probe setting. Further comparisons of orthogonalization strategies are provided in the supplementary material.

Fig.~\ref{fig:mechanism}(b) assesses spatial agreement using the fractions skill score (FSS) across different neighborhood sizes. CorrectCast achieves higher FSS than the base nowcast at all evaluated scales. Removing Wavelet SAFM reduces FSS across these scales, with a more pronounced decrease at larger neighborhood sizes. This pattern suggests that low-frequency wavelet modulation complements spatial adaptation by improving agreement over broader precipitation structures.

Fig.~\ref{fig:mechanism}(c) examines changes in residual power spectral density (PSD) relative to the base nowcast. Positive values indicate reductions in prediction-error power at the corresponding spatial frequencies. CorrectCast reduces residual power across most of the spectrum, with mean decreases of 68.93\%, 25.19\%, and 8.56\% in the low-, mid-, and high-frequency bands, respectively. The greater reduction at low frequencies suggests improved correction of large-scale precipitation errors, consistent with the intended role of Wavelet SAFM.

\begin{figure}[h]
    \centering
    \includegraphics[width=1.0\linewidth]{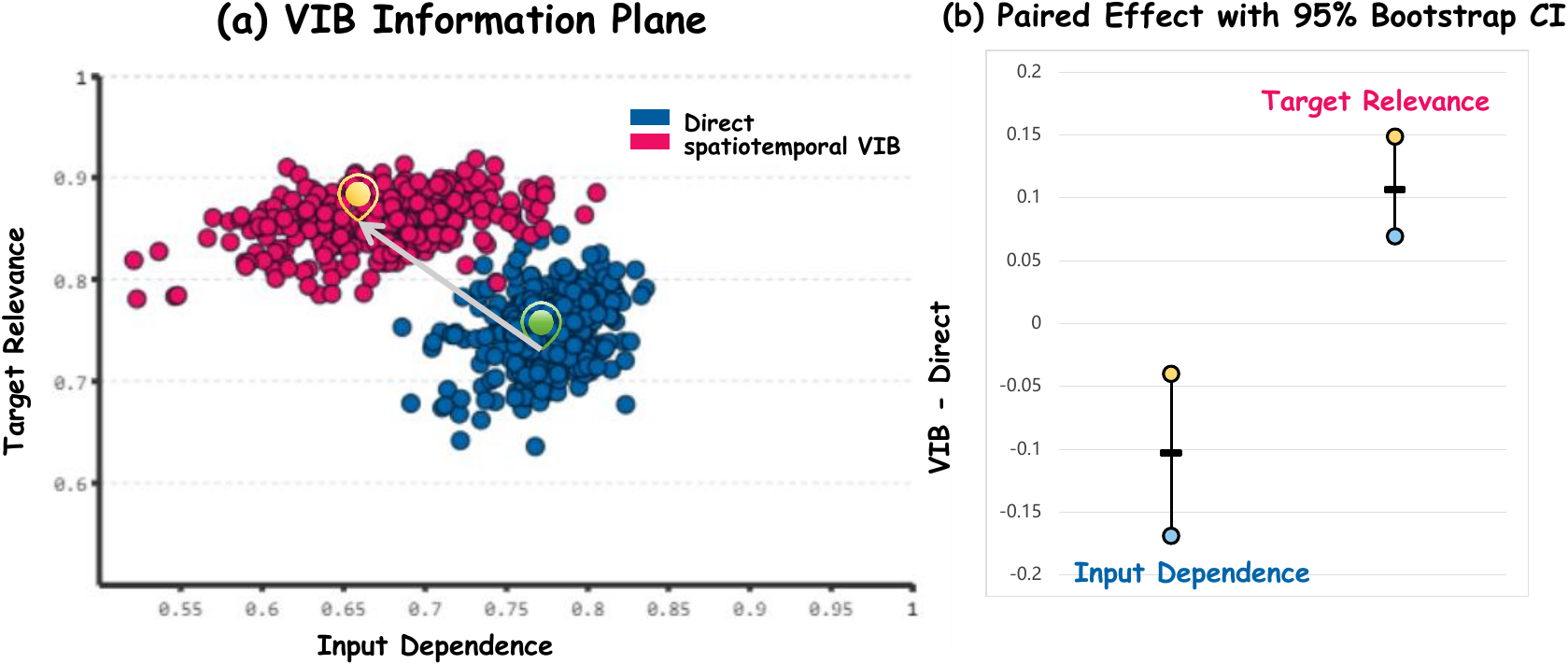}
    \caption{
Information-dependence analysis of the spatiotemporal VIB module.
(a) VIB information plane; preferred direction is lower input dependence with higher target relevance.
(b) Paired deltas over direct fusion (95\% bootstrap CI); $\Delta$input \textless 0 and $\Delta$target \textgreater 0 denote the intended spatiotemporal VIB behavior.
}
    \label{fig:vib}
\end{figure}

We further assess statistical dependence using the normalized Hilbert--Schmidt independence criterion (nHSIC). As shown in Fig.~\ref{fig:vib}, spatiotemporal VIB reduces input dependence from 0.7747 to 0.6718 while increasing target dependence from 0.7496 to 0.8561, relative to direct fusion. The paired differences are $-0.1029$ and $+0.1065$, respectively, and the corresponding 95\% bootstrap confidence intervals both exclude zero. These results indicate a more favorable balance between input compression and target relevance, consistent with the intended role of spatiotemporal VIB in learning correction-oriented representations.
\begin{table}[t]
\centering
\caption{Predictive utility of PWV components under frozen-encoder residual probes. The best results are highlighted in \textbf{bold} and the second-best results are \underline{underlined}.}
\label{tab:orth_probe_main}
\footnotesize
\setlength{\tabcolsep}{4.5pt}
\renewcommand{\arraystretch}{1.1}
\begin{tabular}{l|cccc}
\toprule
Input
& CSI $\uparrow$
& HSS $\uparrow$
& CSI-pool16 $\uparrow$ 
& MSE $\downarrow$\\
\midrule
$\{{Y^{0}, Q}\}$
& 0.1708
& 0.2586
& 0.3915 
& \textbf{34.0149}\\

$\{{Y^{0}, Q, H_{\parallel}^{P}\}}$
& \underline{0.1722}
& \underline{0.2605}
& \underline{0.3917} 
& 34.5913\\

$\{{Y^{0}, Q, H_{\perp}^{P}\}}$
& \textbf{0.1844}
& \textbf{0.2830}
& \textbf{0.4609} 
& \underline{34.1108}\\

$\{{Y^{0}, Q, H_{\parallel}^{P}, H_{\perp}^{P}\}}$
& 0.1673
& 0.2535
& 0.3850 
& 34.2301\\
\bottomrule
\end{tabular}
\end{table}
\subsubsection{Does CorrectCast Preserve Motion Consistency?}
\begin{figure}[h]
    \centering
    \includegraphics[width=1.0\linewidth]{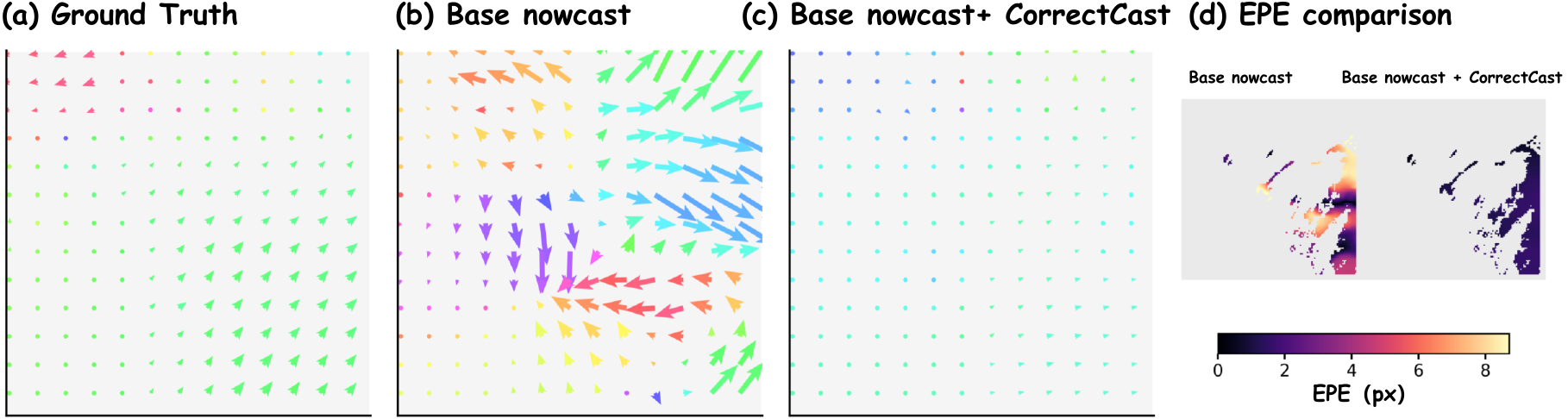}
    \caption{
        (a)-(c) Estimated motion fields from ground-truth, base nowcast, and corrected nowcast (arrows: direction and speed). (d) EPE maps between the base nowcast and the CorrectCast-corrected nowcast.
        }
    \label{fig:motion_consistency}
\end{figure}

We estimate optical flow from consecutive precipitation frames and evaluate motion agreement using endpoint error (EPE) between the base nowcast and the CorrectCast-corrected nowcast. As shown in Fig.~\ref{fig:motion_consistency}, the base nowcast exhibits a fragmented flow pattern with inconsistent motion directions across neighboring regions. In contrast, the corrected nowcast produces a more coherent flow field that better matches the motion estimated from ground-truth precipitation. The corresponding EPE maps show a concentrated region of large errors in the base nowcast, which is substantially reduced after correction. The mean EPE decreases from 4.7610 to 1.3641 pixels, indicating improved agreement with the reference motion field.
\subsubsection{How Well Does CorrectCast Transfer Across Base Nowcasters?}
\begin{figure}[h]
    \centering
    \includegraphics[width=1.0\linewidth]{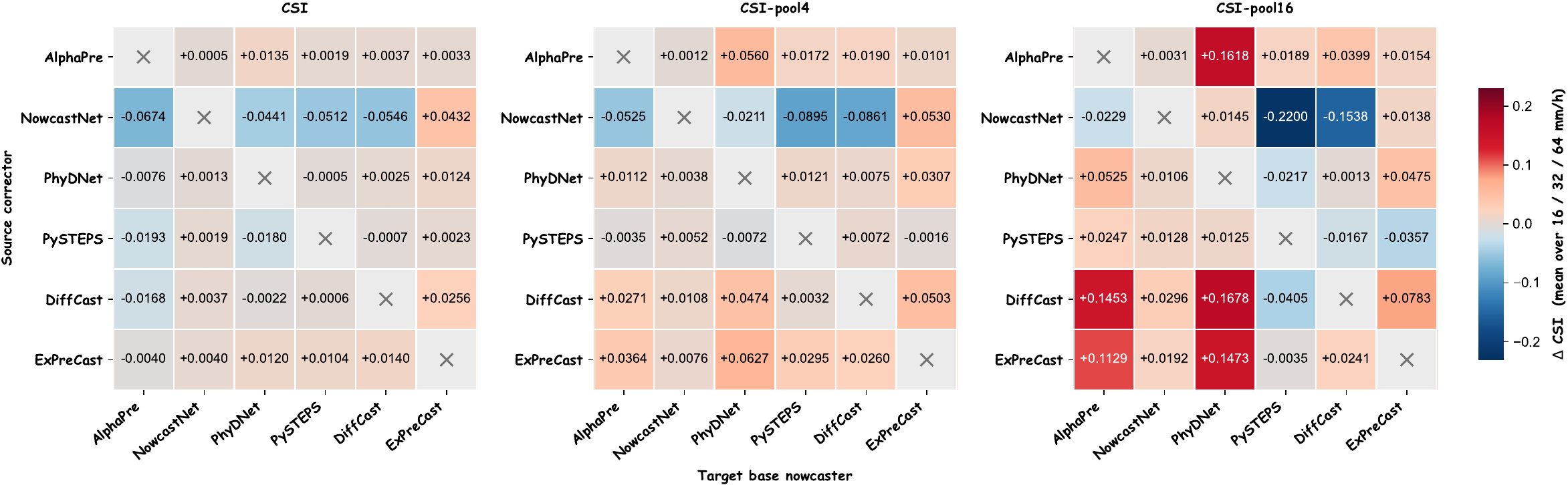}
    \caption{Cross-model transfer on CONUS. X-axis: target base nowcaster receiving the corrector without retraining; Y-axis: source base nowcaster used to train the corrector. Off-diagonal cells report absolute CSI, CSI-pool4, and CSI-pool16 changes over the corresponding target baseline; diagonals omitted.}
    \label{fig:cross_model_transfer}
\end{figure}

We evaluate cross-model transfer on CONUS by applying correctors trained with one base nowcaster to other nowcasters without retraining. Fig.~\ref{fig:cross_model_transfer} shows that improvements are less consistent than in the matched-model setting, with positive gains occurring more frequently for spatially pooled CSI than for pixelwise CSI. Notably, the AlphaPre-trained corrector improves all three CSI metrics for every other base nowcaster, whereas the NowcastNet-trained corrector degrades performance for several target models. These results show that CorrectCast can be applied across different nowcasting architectures without modification, although successful transfer depends on the source and target models.

\subsubsection{How Well Does CorrectCast Transfer from MRB to CONUS?}
\begin{figure}[t]
    \centering
    \includegraphics[width=1.0\linewidth]{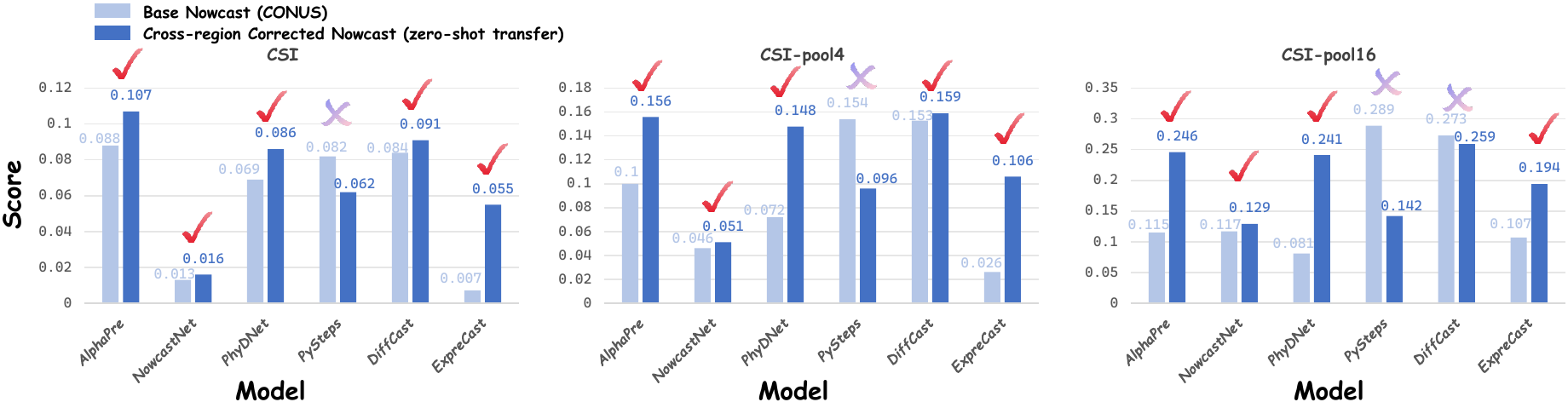}
    \caption{Cross-region transfer from MRB to CONUS. The base nowcast and cross-region corrected nowcast are compared using CSI, CSI-pool4, and CSI-pool16.
}
    \label{fig:cross_region_transfer}
\end{figure}

To evaluate cross-dataset transfer, we train CorrectCast on MRB and apply it to CONUS without retraining. As shown in Fig.~\ref{fig:cross_region_transfer}, CSI improves for five of the six base nowcasters, with gains in spatially pooled CSI also observed for AlphaPre, NowcastNet, PhyDNet, and ExPreCast. However, PySTEPS exhibits decreases across all three CSI metrics, while DiffCast shows a slight decline in CSI-pool16. These mixed results suggest partial transferability between the two datasets, while indicating that the effectiveness of transferred corrections varies with the base nowcaster and target-domain conditions.
\begin{table}[t]
    \centering
    \caption{Computational efficiency comparison under the same evaluation environment. The CorrectCast row reports only the additional overhead introduced by the correction module.}
    \label{tab:efficiency_comparison}
    \setlength{\tabcolsep}{1pt}
    \renewcommand{\arraystretch}{1.0}
    \resizebox{\columnwidth}{!}{
    \begin{tabular}{@{}lrrrrr@{}}
        \toprule
        & Params(M)$\downarrow$
        & \multicolumn{2}{c}{Training}
        & \multicolumn{2}{c}{Inference} \\
        \cmidrule(lr){3-4}
        \cmidrule(lr){5-6}
        Models
        & 
        & Time(h)$\downarrow$
        & Mem.(MB)$\downarrow$
        & Latency(ms)$\downarrow$
        & Mem.(MB)$\downarrow$ \\
        \midrule
        PySTEPS~\cite{pysteps}
        & --- & --- & --- & 3159.02 & --- \\
        PhyDNet~\cite{phydnet}
        & 3.092 & 35.50 & 4876 & 1480.30 & 26 \\
        AlphaPre~\cite{alphaPre}
        & 89.026 & 35.75 & 35555 & 1092.53 & 1363 \\
        ExPreCast~\cite{exprecast}
        & 32.005 & 25.40 & 7217 & 742.37 & 212 \\
        NowcastNet~\cite{nowcastnet}
        & 34.932 & 29.46 & 5050 & 1831.36 & 10008 \\
        DiffCast~\cite{diffcast}
        & 49.362 & 50.75 & 23059 & 10356.23 & 365 \\
        \midrule
        \textbf{CorrectCast}
        & 0.985
        & 14.27
        & 12389
        & 70.29
        & 184 \\
        \bottomrule
    \end{tabular}
    }
\end{table}
\subsubsection{How Much Computational Overhead Does CorrectCast Introduce?}
Table~\ref{tab:efficiency_comparison} reports the computational costs of different nowcasting models. CorrectCast trains only its correction module, requiring 14.27 hours compared with 25.40--50.75 hours for the evaluated deep-learning models. It introduces only 0.985 M trainable parameters, with 70.29 ms inference latency and 184 MB peak GPU memory. These results demonstrate that CorrectCast offers an efficient alternative to full-model retraining when existing nowcasting models underperform. PySTEPS is a non-learning extrapolation method and is excluded from the training-cost comparison.
\section{Conclusion}
We presented CorrectCast, a decoupled residual correction framework that refines completed nowcasts using historical QPE and PWV without updating base nowcasters. Experiments across six nowcasters on MRB and CONUS demonstrate consistent improvements in categorical skill, spatially pooled CSI, and image quality. Ablation and mechanism analyses support the proposed multimodal context selection and scale-aware correction, while transfer experiments reveal both generalization potential and limitations. Overall, CorrectCast offers a lightweight approach to improving existing nowcasts without retraining the base nowcaster.

\section*{LLM Usage Statement}
The authors used ChatGPT(OpenAI) solely to assist in refining the language, improving readability, and ensuring clarity in various sections of the paper. It is important to note that the LLM was not involved in the ideation, research methodology, or experimental design. All research concepts, ideas, and analyses were developed and conducted by the authors. The authors take full responsibility for the content of the manuscript. We have ensured that the LLM-generated text adheres to ethical guidelines and does not contribute to plagiarism or scientific misconduct.
{
    \small
    \bibliographystyle{ieeenat_fullname}
    \bibliography{main}
}

% WARNING: do not forget to delete the supplementary pages from your submission 
% \input{sec/X_suppl}

\end{document}